\documentclass[conference]{IEEEtran}
\IEEEoverridecommandlockouts
\usepackage{times}
\usepackage{latexsym}
\usepackage{amsmath,amssymb,amsfonts}
\usepackage{algorithmic}
\usepackage{graphicx}
\usepackage{textcomp}
\usepackage{bbm}
\usepackage{booktabs}
\usepackage{hyperref}
\hypersetup{hidelinks}
\usepackage{cite}
\usepackage{xcolor}
\usepackage{placeins}
\providecommand{\BIBentryALTinterwordspacing}{\spaceskip=\fontdimen2\font plus 1fil minus \fontdimen4\font\relax}

\def\BibTeX{{\rm B\kern-.05em{\sc i\kern-.025em b}\kern-.08em
    T\kern-.1667em\lower.7ex\hbox{E}\kern-.125emX}}
\begin{document}

\title{Evolving Knowledge Distillation for Lightweight Neural Machine Translation
\thanks{Accepted for publication in the Proceedings of the 2025 IEEE International Conference on Tools with Artificial Intelligence (ICTAI 2025). DOI: 10.1109/ICTAI66417.2025.00092 \copyright\ 2025 IEEE. Personal use of this material is permitted. Permission from IEEE must be obtained for all other uses, in any current or future media, including reprinting/republishing this material for advertising or promotional purposes, creating new collective works, for resale or redistribution to servers or lists, or reuse of any copyrighted component of this work in other works.}
}


\author{\IEEEauthorblockN{1\textsuperscript{st} Xuewen Zhang}
\IEEEauthorblockA{\textit{Department of Content Generation} \\
\textit{Li Auto }\\
Beijing, China\\
xuewenpek@pku.org.cn}
\and
\IEEEauthorblockN{2\textsuperscript{nd} Haixiao Zhang}
\IEEEauthorblockA{\textit{Department of Content Generation} \\
\textit{Li Auto}\\
Beijing, China \\
zhanghaixiao@lixiang.com}
\and
\IEEEauthorblockN{3\textsuperscript{rd} Xinlong Huang}
\IEEEauthorblockA{\textit{Department of Content Generation} \\
\textit{Li Auto}\\
Beijing, China \\
huangxinlong@lixiang.com}

}

\maketitle

\begin{abstract}
Recent advancements in Neural Machine Translation (NMT) have significantly improved translation quality. However, the increasing size and complexity of state-of-the-art models present significant challenges for deployment on resource-limited devices. Knowledge distillation (KD) is a promising approach for compressing models, but its effectiveness diminishes when there is a large capacity gap between teacher and student models. To address this issue, we propose Evolving Knowledge Distillation (EKD), a progressive training framework in which the student model learns from a sequence of teachers with gradually increasing capacities. Experiments on IWSLT-14, WMT-17, and WMT-23 benchmarks show that EKD leads to consistent improvements at each stage. On IWSLT-14, the final student achieves a BLEU score of 34.24, narrowing the gap to the strongest teacher (34.32 BLEU) to just 0.08 BLEU. Similar trends are observed on other datasets. These results demonstrate that EKD effectively bridges the capacity gap, enabling compact models to achieve performance close to that of much larger teacher models.Code and models are available at https://github.com/agi-content-generation/EKD.
\end{abstract}



\section{Introduction}
\begin{figure*}[h]
    \centering
    \includegraphics[width=\linewidth]{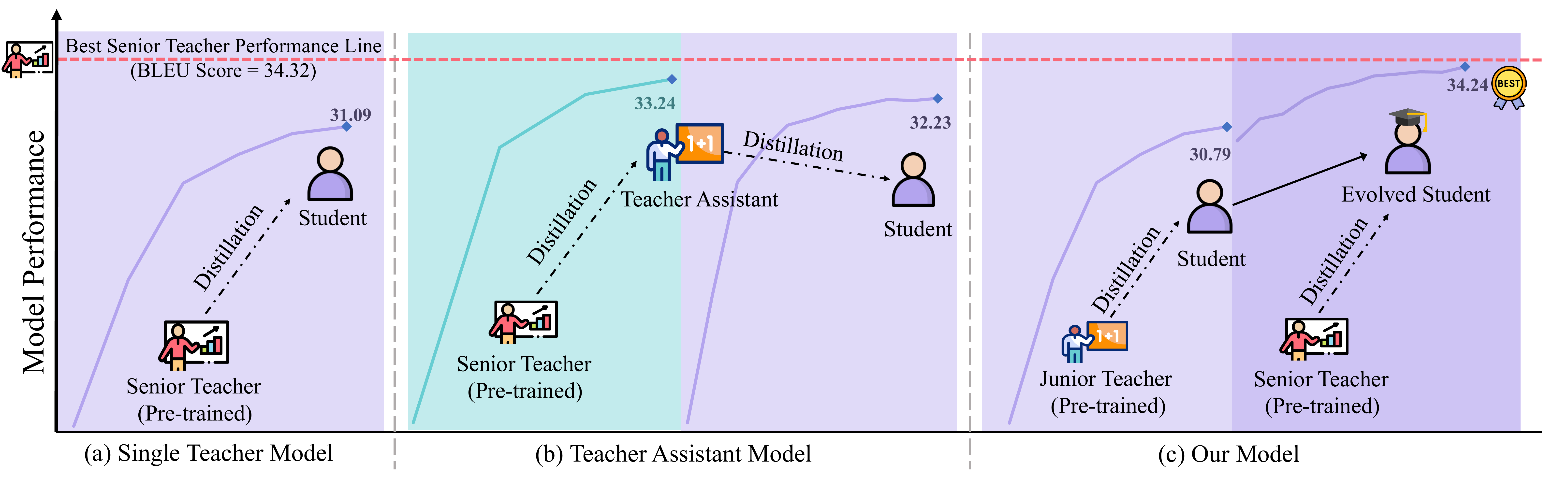}
    \caption{
    Model Performance Across Different Distillation Processes. The red dashed line marks the senior teacher's best BLEU score (34.32, higher is better). Purple and green lines track model performance during distillation, with blue diamonds showing each model's best result. (a) Single Teacher: Direct distillation from the senior teacher shows the largest performance gap. (b) Teacher Assistant: Using an intermediate assistant teacher narrows the gap. (c) Our Model: Progressive distillation from a junior to a senior teacher achieves the closest performance to the senior teacher.}
    \label{fig: ourmethod}
\end{figure*}

Neural Machine Translation (NMT) has emerged as a powerful end-to-end approach for automated translation, employing a single neural network to directly model the probability of a target sentence given a source sentence \cite{bahdanau2015neural}. In recent years, NMT models have significantly improved translation quality, accompanied by a substantial expansion in model scale. 
Since Transformer introduced \cite{vaswani2017attention}, the parameter count of NMT models has grown exponentially. For instance, M2M-100 (12 billion parameters) \cite{fan2021beyond} and NLLB (54 billion parameters) \cite{koishekenov2023memory} have grown dramatically in size, supporting translation across hundreds of languages.
However, these large-scale models also incur significant computational and storage costs, limiting their deployment on resource-constrained devices. Consequently, the question of how to reduce model size while maintaining translation performance has become an urgent issue. In this context, knowledge distillation \cite{geoffrey2014vo, gou2021knowledge, li2024continual}, as an effective method for model compression, has gained increasing attention in the NMT.

While knowledge distillation has shown promise in compressing large pre-trained NMT models, recent studies have demonstrated that when the capacity gap between teacher and student models is too large, the student struggles to mimic the teacher \cite{wei2024sentence, zhang2023lifting}, as illustrated in Fig. \ref{fig: ourmethod} (a). This leads to suboptimal performance and limits the effectiveness of this technique \cite{cho2019efficacy}. This disparity not only hinders the student's ability to capture the teacher's knowledge fully but also constrains the teacher's potential to guide the student effectively \cite{mirzadeh2020improved}. Therefore, it is crucial to find an effective distillation method that maximizes the retention of the teacher model's abilities in the student model.

The Teacher-Assistant Knowledge Distillation (TAKD) framework has been proposed as a potential way to address the capacity gap challenge \cite{mirzadeh2020improved, han2025rethinking}. This approach introduces a medium-sized teacher assistant (TA) model, which has a capacity between the large teacher and the small student. The TA first learns from the teacher and then guides the student, as illustrated in Fig. \ref{fig: ourmethod} (b). Although this hierarchical distillation method enhances the student's learning effectiveness compared to a single-teacher approach, significant performance degradation occurs at each distillation stage
\cite{cho2019efficacy}. Consequently, the resulting student model often falls short of fully capturing the teacher's performance. This persistent performance gap necessitates more effective techniques to retain the teacher model's capabilities in the student model.

To address these limitations, we propose a novel evolving learning strategy called \textbf{E}volving \textbf{K}nowledge \textbf{D}istillation (\textbf{EKD}). Our approach mimics human cognitive development, where students learn from teachers matched to their stage, from primary school to university. Through this process, the student models gradually enhance their learning capabilities and knowledge base.
Specifically, our method introduces a single student model that learns from a series of teacher models with increasing capacities (i.e., the number of parameters), see Fig. \ref{fig: ourmethod} (c). This approach allows the student model's ability to continuously evolve and more effectively bridge the capacity gap.

Our experimental results demonstrate the efficacy of our proposed method. As the student model progresses through learning with increasingly capable teachers, we observe a consistent improvement in its translation performance. Our approach yields superior results compared to the single teacher model, with the student model more closely approximating the performance of the large teacher model, effectively addressing the capacity gap problem. Furthermore, in contrast to the teacher-assistant method, which often suffers from performance degradation in successive stages, our EKD enables the student model to enhance its capabilities incrementally. This stepwise improvement not only results in performance more comparable to that of a large teacher model but also equips the student model with the ability to adapt to and learn from newly introduced, larger teacher models.

Our contributions can be summarized in the following three aspects:
\begin{itemize}
\item\textbf{Novel Progressive Learning Strategy}: We introduce a novel progressive learning strategy for knowledge distillation in NMT, where a single student model learns sequentially from a series of teacher models with increasing capacities. 

\item\textbf{Effective Capacity Gap Resolution}: Our method allows the student model to more closely approximate the performance of the full-scale teacher model, demonstrating superior performance in resolving the capacity gap problem.

\item\textbf{Improved Performance Retention}: Unlike the teacher-assistant method, which often suffers from performance degradation in successive distillation stages, our approach continuously enhances the student model's ability, enabling it to learn from increasingly larger teacher models.
\end{itemize}

\section{Related work}

\noindent\textbf{Single Teacher Knowledge Distillation}.
Single Teacher Knowledge Distillation has been extensively explored as a means to leveraging the knowledge of a larger, more complex teacher model to improve a smaller student model \cite{10097992, fang2025knowledge}. 
In the field of NMT, single-teacher distillation has also shown promising results and can be categorized into sentence-level and token-level approaches based on their training targets \cite{jung2023feature, wei2024sentence, galiano2025beyond}. 
Sentence-level distillation uses teacher-generated pseudo-targets to simplify learning for the student model, focusing on overall translation accuracy \cite{kim2016sequence,tang2021vidlankd, tan2022document}. 
Token-level distillation, operating at a finer granularity, trains the student to mimic the teacher's token-level distributions, capturing nuanced differences and improving performance on lexically diverse knowledge \cite{wang2020structure, wang2021selective, zhang2025aligndistil}. 

\noindent\textbf{Teacher Assistant Knowledge Distillation}. 
Recent advances in knowledge distillation have introduced Teacher Assistant Knowledge Distillation (TAKD) as a promising method to bridge the gap between teacher and student models \cite{furlanello2018born, mirzadeh2020improved, jia2024distillsleepnet}. 
TAKD employs an intermediate teacher assistant, distilled from the teacher model, to facilitate knowledge transfer to the student. Subsequent research has focused on optimizing the selection of teacher assistants (TAs). Reference \cite{son2021densely} proposed using multiple TAs with gradually decreasing model sizes, while reference \cite{han2024amd} developed an approach to automatically identify the optimal TA by balancing performance and scale in large-scale vision model compression. In the context of NMT, reference \cite{lv2024taekd} introduced TAeKD, which designs a fusion model integrating translation outputs from multiple closed-source models as a teacher assistant to generate soft labels and training samples. 

\noindent\textbf{Multi-Teacher Knowledge Distillation}. 
Multi-Teacher Knowledge Distillation can be categorized into two main approaches: Simultaneous and Progressive. In Simultaneous Knowledge Distillation, all teachers are required to be available at the same time. For instance, reference \cite{tan2019multilingual, thrivikram2021efficient} proposed methods where a multilingual student model simultaneously aligns outputs from teacher models with different language knowledge. Reference \cite{saleh2020collective,zhang2023adaptive} developed adaptive distillation methods for training student models from teacher ensembles. 
However, Simultaneous Knowledge Distillation can lead to substantial memory consumption as the number of teachers increases and limits the incorporation of newly released teachers.

In contrast, our work is based on Multi-Teacher Progressive Knowledge Distillation, which enables sequential learning from multiple teachers.
Reference \cite{cao2023learning} proposed that the learning sequence should be based on the architectural similarity between teacher-student pairs, facilitating distillation across different architectures. This method successfully transfers knowledge from transformer-based teachers to convolution-based students. 
To address the problem of knowledge overwriting caused by different domains of teachers, reference \cite{zhang2023towards} filters new knowledge from the current teacher while preserving valuable information from previous iterations.

\section{Method}

In this section, we present our EKD method for NMT tasks. 
We structure our methodology as follows: First, in Section \ref{sec: NMT}, we outline the training objective for the NMT task. Section \ref{sec: single} then details the distillation process from a single teacher to a single student. Finally, in Section \ref{sec: progressive}, 
we elucidate the specific steps of evolving distillation, demonstrating how a small student model evolves by learning from increasingly larger teacher models, thereby gradually enhancing its capabilities. 
In this work, we focus on a two-teacher distillation approach to demonstrate the effectiveness of our method, while noting that it can be readily extended to scenarios involving multiple teachers.

\subsection{Neural Machine Translation}\label{sec: NMT}
Let $V$ represent the vocabulary of the target language, with $|V|$ indicating its size. Given a source sentence $\mathbf{x} = (x_1, \ldots, x_n)$ of length $n$ and a target sentence $\mathbf{y} = (y_1, \ldots, y_m)$ of length $m$, where $y_t \in V$ for all $t$, our objective is to maximize the probability of correctly predicting each target word. The training objective for token-level NMT is formulated as the minimization of the cross-entropy loss function, expressed as:
\begin{equation}
  \mathcal{L}_{ce}(\theta) = - \sum_{t=1}^m\sum_{k=1}^{|V|}\mathbbm{1}\{y_t = k\}
  \log p_\theta(y_t=k|\mathbf{y}_{<t},\mathbf{x})  
  \label{eq:nmt_loss}
\end{equation}
where $\mathbbm{1}\{y_t = k\}$ is an indicator function that equals 1 if the current target word $y_t$ is the $k$-th word in the vocabulary $V$, and 0 otherwise. The term $p_\theta(y_t=k|\mathbf{y}_{<t},\mathbf{x})$ represents the model's predicted probability that the $t$-th word is the $k$-th word in the vocabulary, conditioned on the source sentence $\mathbf{x}$ and the preceding target words $\mathbf{y}_{<t}$. Here, $\theta$ denotes the model parameters. This formulation encapsulates an autoregressive process, where each target word is predicted based on the source sentence and all previously generated target words, thereby capturing the sequential dependencies in the translation task.


\subsection{Single Teacher Distillation}\label{sec: single}
In knowledge distillation from a single teacher to a single student, the student model $S$ needs to mimic the predicted probability of the pre-trained teacher model $T$. This is achieved by minimizing the cross-entropy loss using $P_{\theta_T}$ as soft labels, 
\begin{align}
    \mathcal{L}_{kd}(\theta_S|\theta_T) = &-\sum_{t=1}^m\sum_{k=1}^{|V|}
    P_{\theta_T}(y_t=k|\mathbf{y}_{<t},\mathbf{x})\times \notag\\
    &\log P_{\theta_S}(y_t=k|\mathbf{y}_{<t},\mathbf{x}) 
\end{align}
where $\theta_T, \theta_S$ denote the model parameters for the teacher model and student model, respectively.
Meanwhile, the student model should learn to accurately predict target words by minimizing the cross-entropy loss $\mathcal{L}_{ce}(\theta_S)$. Overall,
the objective function is:
\begin{align}
    \mathcal{L}(\theta_S|\theta_T) = \mathcal{L}_{ce}(\theta_S)+\alpha \mathcal{L}_{kd}(\theta_S|\theta_T)
\end{align}
where $\alpha$ is a hyper-parameter to balance two losses.

\subsection{Evolving Distillation}\label{sec: progressive}
To bridge the gap between the small student model $S$ and the large teacher model $T$, we propose an evolving learning approach where the student model learns sequentially from teacher models of increasing capacities.

In this paper, we work under the homogeneous teacher assumption, where the student and teacher models share the same architecture (i.e., transformer-based) but differ in parameter count. This approach enhances the student model's potential to continuously evolve its learning ability by gradually learning from capacity-comparable teachers.

\subsubsection{Teacher Model Hierarchy}

We establish a hierarchical structure of teacher models with increasing capacities to guide the progressive knowledge transfer process. We call the larger teacher model as the \textit{senior teacher} model, denoted as $T_{\textit{senior}}$ and the smaller teacher model as the \textit{junior teacher} model, denoted as $T_{\textit{junior}}$. The parameter count of the junior teacher is intermediate between the student model and the senior teacher, i.e., $N_S<N_{T_{\textit{junior}}}<N_{T_{\textit{senior}}}$.

We postulate that the learning dynamics can be represented as a function of the relative capacity difference:

\begin{align}
\Delta_{\text{learn}}(M_1, M_2) \propto f(\frac{N_{M_1} - N_{M_2}}{N_{M_1}})
\end{align}

where $\Delta_{\text{learn}}$ represents the effectiveness of knowledge transfer between models $M_1$ and $M_2$, and $f$ is a monotonically decreasing function indicating that learning becomes more challenging as the capacity gap increases.

This hierarchical arrangement creates a smoother knowledge transition pathway where each step involves a more manageable capacity gap.

\subsubsection{Two-Stage Distillation Process}

The distillation process occurs in two stages designed to optimize the knowledge transfer pathway.

\paragraph{Stage 1: Junior Teacher Distillation}
First, the student model $S$ is distilled from the more comparable pre-trained junior teacher model $T_{\textit{junior}}$. We denote this process as $T_{\textit{junior}}\rightarrow{S}$, with the corresponding loss function:
\begin{align}
\mathcal{L}_{T_{\textit{junior}}\rightarrow{S}}(\theta_S|\theta_{T_{\textit{junior}}}) = \alpha \mathcal{L}_{\text{KL}}(S, T_{\textit{junior}}) + 
\notag \\
(1-\alpha)\mathcal{L}_{\text{task}}(S)
\end{align}

where $\mathcal{L}_{\text{KL}}$ represents the Kullback-Leibler divergence between the output distributions of the student and junior teacher models, $\mathcal{L}_{\text{task}}$ is the task-specific loss function, and $\alpha$ is a hyperparameter controlling the trade-off between mimicking the teacher and optimizing for the task directly.

\paragraph{Stage 2: Senior Teacher Refinement}
Subsequently, the student model continues its learning under the guidance of the senior teacher, denoted as $T_{\textit{senior}}\rightarrow{[T_{\textit{junior}}\rightarrow{S}]}$. Using $\theta^{'}_S$ to denote the evolved student model parameters after the first stage, the loss function for this stage is:
\begin{align}
\mathcal{L}_{T_{\textit{senior}}\rightarrow{[T_{\textit{junior}}\rightarrow{S}]}}(\theta^{'}_S|\theta_{T_{\textit{senior}}}) = &\beta \mathcal{L}_{\text{KL}}(S', T_{\textit{senior}}) + 
\notag \\
&(1-\beta)\mathcal{L}_{\text{task}}(S')
\end{align}

where $\beta$ may differ from $\alpha$ to reflect the changing importance of teacher guidance as training progresses.

\subsubsection{Curriculum Learning Perspective}

Our evolving distillation approach can be viewed through the lens of curriculum learning, where the complexity of the learning target gradually increases. The junior teacher provides a more attainable initial target that serves as a stepping stone toward capturing the capabilities of the senior teacher.

Formally, we hypothesize that the knowledge state of the student model after the two-stage process is superior to that achievable through direct distillation from the senior teacher:

\begin{align}
\mathcal{K}(T_{\textit{senior}}\rightarrow{[T_{\textit{junior}}\rightarrow{S}]}) > \mathcal{K}(T_{\textit{senior}}\rightarrow{S})
\end{align}

where $\mathcal{K}(\cdot)$ represents a measure of the knowledge successfully transferred to the student model.

\subsubsection{Continuous Evolution Path}
While we demonstrate our approach with a two-teacher setup, our evolving distillation framework naturally extends to scenarios with multiple teachers arranged in a capacity-ascending sequence:
\begin{align}
T_1 \rightarrow T_2 \rightarrow \cdots \rightarrow T_n \rightarrow S
\end{align}

This continuous evolution path allows for even more gradual knowledge transfer, with each intermediate teacher further refining the student's capabilities. The number of teachers and their respective capacities can be tailored to specific deployment constraints and performance requirements.

This evolving distillation process allows for a more gradual and effective transfer of knowledge from larger models to smaller ones, enabling compact models to achieve superior performance compared to traditional single-teacher distillation methods.

\section{Experiments}

\subsection{Experimental Setups}

\noindent\textbf{Datasets}
We evaluate our proposed method on three public datasets: German-English (De-En), English-Czech (En-Cs), and English-German (En-De). The details of each data are as follows:

\begin{itemize}
     \item For the De-En task, we utilize the iwslt-14-de-en corpus, which consists of TED talk transcripts. The training set comprises 160k sentence pairs, while the validation and test sets contain 7k and 6k sentence pairs, respectively.
    \item For the En-Cs task, we employ the wmt-23-en-cs dataset, which is derived from NLP paper abstracts. The training set contains 247k sentence pairs, with 3k pairs in the validation set and 1k pairs in the test set.
    \item For the En-De task, we use the wmt-17-en-de, which includes 258k sentence pairs for training. Both the validation and test sets consist of 3k sentence pairs each.
\end{itemize} 

\begin{table}[!t]
    \centering
    \caption{Model parameter sizes for three datasets.}
    \renewcommand{\arraystretch}{1.2}
    \resizebox{1\linewidth}{!}{
    \begin{tabular}{lccc}
    \toprule
        \textbf{Models} & \textbf{iwslt-14-de-en} & \textbf{wmt-23-en-cs} & \textbf{wmt-17-en-de} \\ \midrule
        $S$ & 6M & 15M & 14M \\ 
        $T_{\textit{junior}}$ & 15M & 31M & 31M \\ 
        $T_{\textit{senior}}$ & 39M & 72M & 71M \\ \bottomrule
    \end{tabular}
    }
    \label{tab:model_parameters}
\end{table}

All the sentences are first tokenized with moses tokenizer\footnote{\url{https://github.com/moses-smt/mosesdecoder/blob/master/scripts/tokenizer/tokenizer.perl}} and then segmented into subword symbols using Byte Pair Encoding (BPE) \cite{sennrich2016neural}. We perform the BPE merge operations across all the languages and keep consistent vocabulary between the teacher and student models to facilitate knowledge distillation.

\noindent\textbf{Model Configuration}
In our study, we employ the transformer\cite{vaswani2017attention}  as the foundational structure for our model. This choice is motivated by the need for the research model to evolve from a base model, in addition to which most advanced models today are based on the Transformer architecture.
Additionally, to facilitate effectiveness of the models evolution and enable each hierarchical level to fully learn from the capabilities of the upper-level models,  we maintained the parameter counts of each model within a factor of 2 to 3 times that of its preceding level \cite{zhou2019understanding, zhang2023lifting, wei2024sentence, peng2024pre}. The number of parameters for each model is shown in Table \ref{tab:model_parameters}.

We utilize three models with varying configurations: For the student model $S$, the parameters are set to an embedding dimension of $128$, a feedforward neural network embedding dimension of $1024, 4$ attention heads, and $6$ layers. The junior teacher model $T_{\textit{junior}}$ has corresponding parameters of $256, 1024, 4,$ and $6$, respectively. Similarly, the senior teacher model $T_{\textit{senior}}$ is configured with parameters of $512$ for the embedding dimension, $1024$ for the feedforward neural network embedding dimension, $4$ attention heads, and $6$ layers.

\noindent\textbf{Training Settings}
We train models on NVIDIA Tesla A100 GPU via Fairseq \cite{ott2019fairseq} toolkit. 
Optimization is performed using the Adam optimizer with $\beta$ parameters set to $(0.9, 0.98)$ and an initial learning rate of $5 \times 10^{-4}$. To gradually increase the learning rate, we employ an inverse square root learning rate scheduler with $4000$ warm-up steps. 
To prevent overfitting, we apply a dropout probability of $0.3$, randomly deactivating neurons during training. Additionally, we use $L2$ regularization with a weight decay coefficient of $0.0001$. 
For the loss function, we utilize label-smoothed cross-entropy with a label smoothing factor of $0.1$ to enhance the model's generalization ability. 
Each training batch contains a maximum of 4096 tokens.
During inference, we employ beam search decoding with a beam size of $5$, processing $128$ sentences at a time. To obtain complete texts, the final translation outputs have the BPE segmentation markers removed. 

\noindent\textbf{Computation Cost}
We summarize the computation cost associated with each model configuration in Table~\ref{tab:newformat}, including total training cost (measured in TotalFLOPs) and the mode of training (training from scratch, denoted as "S", or training from a checkpoint, denoted as "C"). 
In particular, $T_{\textit{senior}}\to [T_{\textit{junior}}\to S]$, which is trained by EKD, does not incur a significant computational overhead. Instead, its cost (0.24 PFLOPs) is even lower than the computational cost required to train the junior teacher alone (0.29 PFLOPs).
\begin{table}[tp]
  \centering
  \caption{Computational Cost (Total FLOPs) of Models}
  \renewcommand{\arraystretch}{1.2}
  \resizebox{0.7\linewidth}{!}{%
    \begin{tabular}{lcc}
      \toprule
      \textbf{Models} &
      \textbf{TotalFLOPs (P)} &
      \textbf{Mode} \\
      \midrule
      $S$                                                        & 0.12 & S \\
      $T_{\textit{junior}}$                                      & 0.29 & S \\
      $T_{\textit{senior}}$                                      & 0.76 & S \\
      $T_{\textit{junior}} \to S$                                & 0.12 & S \\
      $T_{\textit{senior}} \!\!\to\!\! [T_{\textit{junior}} \!\! \to\!\! S]$ & 0.24 & C \\
      \bottomrule
    \end{tabular}%
  }
  \label{tab:newformat}
\end{table}

\noindent\textbf{Evaluation Metrics} 
To evaluate the translation quality, we employ two metrics:
\begin{itemize}
    \item \textbf{BLEU} \cite{papineni2002bleu}, the standard metric for machine translation evaluation. BLEU primarily calculates precision by comparing n-gram matches between the system output and reference translations. We report case-sensitive, detokenized BLEU scores using SacreBLEU\footnote{\url{https://github.com/mjpost/sacrebleu}}.
    \item \textbf{COMET} \cite{rei2020comet}, a neural metric that demonstrates higher correlation with human judgments. COMET provides a more comprehensive assessment of translation quality through semantic and syntactic analysis of the system output. We utilize the publicly available wmt22-comet-da\footnote{\url{https://github.com/Unbabel/COMET}} model to compute COMET scores.
\end{itemize}
\FloatBarrier


\begin{table}[tp]
  \centering
  \caption{BLEU and COMET scores across different models.}
  \renewcommand{\arraystretch}{1.2}
  \resizebox{\linewidth}{!}{%
    \begin{tabular}{lcccccc}
      \toprule
      {\textbf{Models}} &
      \multicolumn{2}{c}{\textbf{iwslt-14-de-en}} &
      \multicolumn{2}{c}{\textbf{wmt-23-en-cs}} &
      \multicolumn{2}{c}{\textbf{wmt-17-en-de}} \\ 
      \cmidrule(lr){2-3} \cmidrule(lr){4-5} \cmidrule(lr){6-7}
      & BLEU & COMET & BLEU & COMET & BLEU & COMET \\
      \midrule
      $S$                       & 28.54 & 0.71 & 11.51 & 0.44 & 11.46 & 0.42 \\
      $T_{\textit{junior}}$     & 32.78 & 0.75 & 16.00 & 0.50 & 16.90 & 0.47 \\
      $T_{\textit{junior}} \to S$ & 30.79 & 0.73 & 14.10 & 0.48 & 14.76 & 0.45 \\
      $T_{\textit{senior}} \!\!\to\!\! [T_{\textit{junior}} \!\!\to\!\! S]$ 
                               & \textbf{34.24} & \textbf{0.77} & \textbf{16.24} & \textbf{0.51} & \textbf{17.14} & \textbf{0.48} \\
      \bottomrule
    \end{tabular}%
  }
  \label{tab:performance}
\end{table}

\begin{table*}[htbp]
  \centering
  \caption{Comparison of single–step distillation and our method (BLEU).}
  \renewcommand{\arraystretch}{1.2}
  \resizebox{\textwidth}{!}{%
    \begin{tabular}{lccccccccccc}
      \toprule
      \multicolumn{4}{c}{\textbf{Single Teacher Model}} &
      \multicolumn{4}{c}{\textbf{Single Teacher Model}} &
      \multicolumn{4}{c}{\textbf{Our Model}} \\
      \cmidrule(lr){1-4}\cmidrule(lr){5-8}\cmidrule(lr){9-12}
      \textbf{Models} & \textbf{iwslt-14} & \textbf{wmt-23} & \textbf{wmt-17} &
      \textbf{Models} & \textbf{iwslt-14} & \textbf{wmt-23} & \textbf{wmt-17} &
      \textbf{Models} & \textbf{iwslt-14} & \textbf{wmt-23} & \textbf{wmt-17} \\
      & (de-en) & (en-cs) & (en-de) &
      & (de-en) & (en-cs) & (en-de) &
      & (de-en) & (en-cs) & (en-de) \\
      \midrule
      $T_{\textit{junior}}$                & 32.78 & 16.00 & 16.90 &
      $T_{\textit{senior}}$                & 34.32 & 17.00 & 17.79 &
      $T_{\textit{senior}}$                & 34.32 & 17.00 & 17.79 \\
      $T_{\textit{junior}}\!\to\!S$        & 30.79 & 14.10 & 14.76 &
      $T_{\textit{senior}}\!\to\!S$        & 31.09 & 13.96 & 14.83 &
      $T_{\textit{senior}}\!\to\![T_{\textit{junior}}\!\!\to\!S]$
                                           & 34.24 & 16.24 & 17.41 \\
      \midrule
      \textit{Difference}                  &  1.99 &  1.90 &  2.14 &
      \textit{Difference}                  &  3.23 &  3.04 &  2.96 &
      \textit{Difference}                  &  0.08 &  0.76 &  0.38 \\
      (\%) & {\textbf{\color{red}↑ 6.07\%}}& {\textbf{\color{red}↑ 11.00\%}} & {\textbf{\color{red}↑ 12.66\%}} &
      (\%) & {\textbf{\color{red}↑ 9.41\%}} & {\textbf{\color{red}↑ 17.00\%}} & {\textbf{\color{red}↑ 16.63\%}} &
      (\%) & {\textbf{\color{red}↑ 0.23\%}} & {\textbf{\color{red}↑ 4.47\%}} & {\textbf{\color{red}↑ 2.13\%}} \\
      \bottomrule
    \end{tabular}%
  }
  \label{tab:comparison}
\end{table*}

\begin{figure}[htp]
    \centering
    \includegraphics[width=1\linewidth]{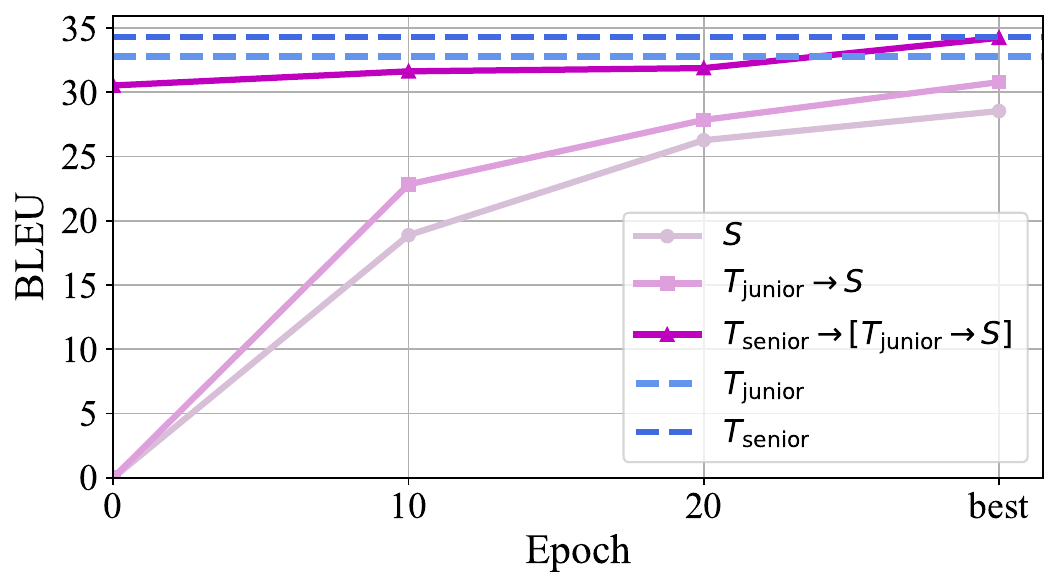} 
    \caption{Model Performance (BLEU) on Iwlst-14-De-En dataset during the training process. Due to varying optimal training durations for different model sizes, the graph displays BLEU scores for the first 20 epochs and the best epoch for clarity.}
    \label{fig:ourmethod}
\end{figure}

\begin{figure*}[h]
    \centering
    \includegraphics[width=1\linewidth]{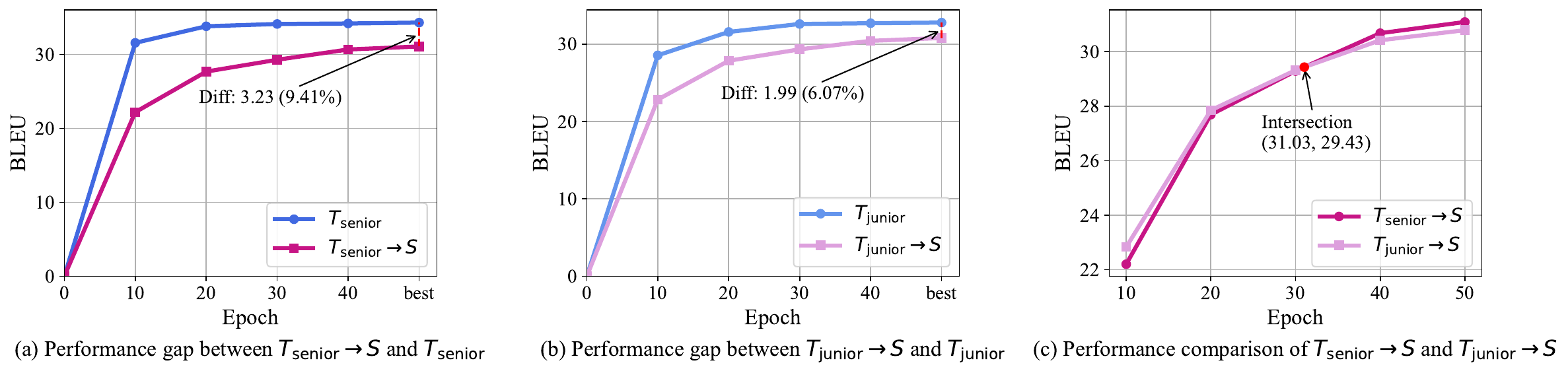}
    \caption{Competence Gap Between Student and Teacher Models and Performance Comparison Among Student Models in Single-Teacher Knowledge Distillation. }
    \label{fig: performance}
\end{figure*}
\subsection{Breaking the learning limits of the model}

In this section, we demonstrate the effectiveness of our proposed EKD method by examining the translation performance throughout the distillation process. Fig. \ref{fig:ourmethod} illustrates the results on the iwlst-14-de-en dataset. The performance trajectory reveals several key insights. When the student model $S$ is distilled from the junior teacher $T_{\textit{junior}}$ (denoted by $T_{\textit{junior}} \rightarrow S$), we observe a significant improvement over the base student model. Subsequently, as the evolved student undergoes further distillation from the senior teacher $T_{\textit{senior}}$ (represented as $T_{\textit{senior}} \rightarrow [T_{\textit{junior} } \rightarrow S]$), the model's performance continues to improve, ultimately approaching the performance of the best senior teacher model. This progression demonstrates the efficacy of our knowledge distillation approach. Notably, the final performance of the student model surpasses that of the best junior teacher model. This phenomenon suggests that our method enables student models to break the original learning limits and outperform their previous teacher counterparts. 

The final results across all three datasets are presented in Table \ref{tab:performance}. This table displays the BLEU and COMET scores of the student model after each distillation step, with the base student and junior model serving as the baseline. Consistently across all datasets, the BLEU and COMET scores of $T_{\textit{senior}} \rightarrow [T_{\textit{junior}} \rightarrow S]$ outperform those of $T_{\textit{junior}} \rightarrow S$ and the base student model $S$, further demonstrating the effectiveness of our distillation approach. Specifically, in terms of BLEU scores, $T_{\textit{junior}} \rightarrow S$ surpasses the base student model $S$ by $|30.79-28.54|/28.54=7.9\%$, $|14.10-11.51|/11.51=22.5\%$ and $|14.76-11.46|/11.46=28.8\%$ on the respective datasets. Furthermore, $T_{\textit{senior}} \rightarrow [T_{\textit{junior}} \rightarrow S]$ outperforms $T_{\textit{junior}} \rightarrow S$ by $11.2\%$, $15.17\%$ and $18.0\%$, respectively. This trend of continuous improvement is also reflected in the COMET metric. In the initial distillation stage, the student model achieves gains of 2.8\%$\sim$9.1\%, followed by additional improvements of 5.5\%$\sim$6.6\% in the second stage. Furthermore, the final distilled student model $T_{\textit{senior}}$ outperforms the base junior model $T_{\textit{junior}}$ across many metrics and datasets. These results provide strong evidence for the progressive evolution of the student model through our proposed distillation process, showcasing substantial performance enhancements at each stage.

To further validate the effectiveness of our proposed method, we extended the hierarchy of teacher models by introducing an additional teacher, denoted as  $T_{master}$. Specifically, after the student model has learned from  $T_{Senior}$, it subsequently learns from  $T_{master}$. Through this multi-stage distillation process, we observe that although the size of the student model remains unchanged, its performance continues to improve. Detailed results are presented in Table \ref{tab:diff_layer_learning}. Notably, the BLEU score of $T_{master}$ reached 34.51, and the student distilled via the complete hierarchy $T_{\textit{master}}\!\!\to\!\!\{T_{\textit{senior}} \!\!\to\!\![T_{\textit{junior}} \!\! \to\!\!  S]\}$ achieved a BLEU score of 34.46. Compared to the original student model, this represents improvements of $7.88\%, 19.7\%$, and $20.74\%$ at different stages. As the number of hierarchical levels increases, the incremental performance improvements gradually slow down, which we attribute to the teacher models approaching their upper capacity bound. Moreover, the performance gap between the student and teacher models narrows significantly, decreasing from an initial $6.07\%$ to just $0.15\%$. This suggests that hierarchical distillation enables the student model to continuously improve and closely approximate the capability of the teacher model. Based on these experimental results, we infer that as long as the teacher models have not reached a performance ceiling, the student model can keep benefiting from increasingly capable teachers, thereby further enhancing its own performance.

\begin{table}[tp]
  \centering
  \caption{Gap between student and teacher models (BLEU).}
  \small
  \renewcommand{\arraystretch}{1.2}
  \resizebox{\linewidth}{!}{%
    \begin{tabular}{lccc}
      \toprule
      \textbf{Models} &
       $\mathbf{T_{\textit{junior}}\!\to\!S}$ &
        $\mathbf{T_{\textit{senior}}\!\to\![T_{\textit{junior}}\!\!\to\!S]}$ &
        $\mathbf{T_{\textit{master}}\!\to\!\{T_{\textit{senior}}\,[T_{\textit{junior}}\!\!\to\!S]\}}$ \\
      \midrule
      Student model  & 30.79 & 34.24 & 34.46 \\
      Teacher model  & 32.78 & 34.32 & 34.51 \\
      \midrule
      \textit{Difference} & 1.99 & 0.08 & 0.05 \\
                   (\%) & {\textbf{\color{red}↑ 6.46\%}} &
                          {\textbf{\color{red}↑ 0.23\%}} &
                          {\textbf{\color{red}↑ 0.15\%}} \\
      \bottomrule
    \end{tabular}%
  }
  \label{tab:diff_layer_learning}
\end{table}

\subsection{Further narrowing the capability gap between student model and teacher model}
In single-teacher knowledge distillation, we observe that the performance gap between the student model and the teacher model increases as the size of the teacher model grows, given a fixed student model. When the student model is distilled from the junior teacher model, the performance gap is 1.99 between $T_{\textit{junior}}$ and $T_{\textit{junior}} \rightarrow S$, accounting for 6.07\% of $T_{\textit{junior}}$'s BLEU score, shown in Fig. \ref{fig: performance} (a). However, when distilled from the senior teacher model, the gap is 3.23 between $T_{\textit{senior}}$ and $T_{\textit{senior}} \rightarrow S$, amounting to approximately 9.41\% of $T_{\textit{senior}}$'s best performance, shown in Fig \ref{fig: performance} (b). Similar trends are observed in the other two datasets. As shown in Table \ref{tab:comparison}, the performance gaps between $T_{\textit{junior}}$ and $T_{\textit{junior}} \rightarrow S$ across the three datasets are 6.07\%, 11\%, and 12.66\%, respectively. In contrast, the gaps between $T_{\textit{senior}}$ and $T_{\textit{senior}} \rightarrow S$ are 9.41\%, 17\%, and 16.63\%, which are on average 4\% larger than those between $T_{\textit{junior}}$ and $T_{\textit{junior}} \rightarrow S$.

This phenomenon can be understood intuitively through human society's experiences: junior high school teachers are generally more suited for teaching junior high school students because their skills and the difficulty of the curriculum align with the abilities of students. Hiring high school teachers to teach junior high school students may lead to counterproductive results. From a modeling perspective, the teacher model's output is too complex for the student model to understand and imitate \cite{cho2019efficacy,mirzadeh2020improved}.



From a deeper perspective, equipping different teachers at various learning stages to maintain a consistently small gap between student and teacher models can enhance the effectiveness of knowledge distillation. Fig. \ref{fig: performance} (c) shows the changes in BLEU scores for $T_{\textit{senior}} \rightarrow S$ and $T_{\textit{junior}} \rightarrow S$ as the training epochs increase.
Initially, $T_{\textit{senior}} \rightarrow S$ performs poorly due to the significantly higher capability of $T_{\textit{senior}}$ compared to $T_{\textit{junior}}$, resulting in increased learning costs during the early stages. However, after 40 epochs of training, $T_{\textit{senior}} \rightarrow S$, having established a foundational understanding, demonstrates a growth trend superior to $T_{\textit{junior}} \rightarrow S$. This is attributed to $T_{\textit{senior}}$'s stronger capabilities compared to $T_{\textit{junior}}$, offering a higher performance ceiling and better matching the current learning ability of $T_{\textit{senior}} \rightarrow S$.

Inspired by these perspectives, in the initial learning phase, we employ $T_{\textit{junior}}$ to distill the student model. Once the student achieves optimal performance at this stage, we switch to using $T_{\textit{senior}}$, which allows the student model to exceed its initial learning limits. In Table \ref{tab:comparison}, $T_{\textit{senior}} \rightarrow [T_{\textit{junior}} \rightarrow S]$ achieves a best BLEU score of 34.24 on the iwslt-14-de-en dataset, just 0.08 lower than $T_{\textit{senior}}$'s best score, representing a minor difference of 0.23\%. Additionally, on the other two datasets, the performance gaps are smaller than those observed with single-teacher distillation, measuring at 4.47\% and 2.13\%, respectively. Our model yields BLEU scores of 34.24, 16.24, and 17.41 for the student model, which is generally about 3 points higher than the scores achieved through single-model distillation (31.09, 13.96, 14.83).

\subsection{Comparison with Teacher Assistant Knowledge Distillation}
We compare our model with the Teacher Assistant Knowledge Distillation (TAKD) method using the iwlst-14-de-en dataset. In this comparison, we use the same teacher model $T_{senior}$. The TAKD method utilizes a medium-scale teacher assistant (TA) to distill knowledge from $T_{senior}$, which is then transferred to the student model. For a fair comparison, we set the size of the TA to be equal to our $T_{junior}$.

\begin{table}[!t]
  \centering
  \caption{Comparison of EKD and TAKD on IWSLT (BLEU).}
  \renewcommand{\arraystretch}{1.2}
  \resizebox{\linewidth}{!}{%
    \begin{tabular}{lclc}
      \toprule
      \multicolumn{2}{c}{\textbf{TAKD}} &
      \multicolumn{2}{c}{\textbf{Our Model (EKD)}} \\
      \cmidrule(lr){1-2} \cmidrule(lr){3-4}
      \textbf{Models} & \textbf{BLEU} & \textbf{Models} & \textbf{BLEU} \\
      \midrule
      Teacher model  & 34.32 & $T_{\textit{senior}}$ & 34.32 \\
      TA             & 33.24 & $T_{\textit{junior}}$ & 32.78 \\
      Student model  & 32.23 &
      $T_{\textit{senior}}\!\!\to\![T_{\textit{junior}}\!\!\to\!S]$ &
      {\textbf{34.24 (\color{red}↑ 5.8\%})} \\
      \bottomrule
    \end{tabular}%
  }
  \label{tab:ta}
\end{table}

As shown in Table \ref{tab:ta}, the BLEU score for the TA in the TAKD method is 33.24, while the final student model achieves a BLEU score of 32.23. In contrast, the final student model of our model, which follows the distillation path $T_{senior} \rightarrow [T_{junior} \rightarrow S]$, achieves a BLEU score of 34.24. This represents an improvement of 5.8\% compared to the student model distilled using the TAKD method.

Furthermore, in the TAKD method, the performance of the student model may not even exceed that of the TA, indicating that the student's performance is limited by the upper bound of the TA. However, our method enables $T_{senior}\rightarrow[T_{junior}\rightarrow S]$ to not only exceed $T_{junior}$ after two distillations but also achieve a BLEU score that is only 0.23\% lower than $T_{senior}$. This highlights the greater potential of our approach. Notably, our student model can further evolve from larger teacher models when compared with TAKD.


\section{Conclusion}
\label{sec:conclusion}
Our work proposes a progressive learning strategy for knowledge distillation in NMT, addressing the performance degradation caused by large capacity gaps between teacher and student models. By leveraging a sequence of teacher models with increasing capacities, our method enables a single student model to progressively enhance its performance, closely approximating that of large-scale teacher models. Experimental results confirm the effectiveness of our approach in achieving high-quality translation with compact models, facilitating efficient deployment on resource-constrained devices.

\section*{Limitations}

We only experimented with a limited set of hyperparameter configurations. While our current settings have demonstrated the effectiveness of our approach, a more thorough tuning of these hyperparameters could potentially yield better performance. However, such extensive hyperparameter optimization is beyond the primary focus of this work.

Furthermore, our experiments have only validated the effectiveness of evolving knowledge distillation when both the teacher and student models are from the same model family (e.g., Transformer-based architectures). It remains an open question how our approach would perform when the student and teacher models adopt heterogeneous architectures (e.g., Transformer-based teachers and Mamba-based student). We leave this as an interesting direction for future work.

\end{document}